\documentclass{article}
\usepackage{spconf,amsmath,epsfig}
\usepackage{booktabs}
\usepackage{enumitem}
\usepackage{color}
\usepackage{amssymb}
\usepackage[table]{xcolor}
\usepackage{appendix}

\definecolor{dkgreen}{rgb}{0,0.6,0}
\definecolor{gray}{rgb}{0.5,0.5,0.5}
\definecolor{mauve}{rgb}{0.58,0,0.82}

\let\OLDthebibliography\thebibliography
\renewcommand\thebibliography[1]{
  \OLDthebibliography{#1}
  \setlength{\parskip}{0pt}
  \setlength{\itemsep}{0pt plus 0.3ex}
}

\def\eg{\emph{e.g., }}

\begin{document}\sloppy
\topmargin=0mm
\def\x{{\mathbf x}}
\def\L{{\cal L}}


	\title{ProTA: Probabilistic Token Aggregation for Text-Video Retrieval}

\name{\parbox{16cm}{\centering
		{\large Han Fang${^*}$\thanks{* Equal contributions.}, Xianghao Zang${^*}$, Chao Ban, Zerun Feng, Lanxiang Zhou, Zhongjiang He, Yongxiang Li, Hao Sun\textsuperscript{$\dag$}\thanks{Corresponding authors.}}}
	\address{China Telecom Corporation Ltd. Data\&AI Technology Company}
}

\maketitle

\begin{abstract}
Text-video retrieval aims to find the most relevant cross-modal samples for a given query.
Recent methods focus on modeling the whole spatial-temporal relations.
However, since video clips contain more diverse content than captions, 
the model aligning these asymmetric video-text pairs has a high risk of retrieving many false positive results.
In this paper, we propose Probabilistic Token Aggregation (\textit{ProTA}) to handle cross-modal interaction with content asymmetry.
Specifically, we propose dual partial-related aggregation to disentangle and re-aggregate token representations in both low-dimension and high-dimension spaces. 
We propose token-based probabilistic alignment to generate token-level probabilistic representation and maintain the feature representation diversity.
In addition, an adaptive contrastive loss is proposed to learn compact cross-modal distribution space.
Based on extensive experiments, \textit{ProTA} achieves significant improvements on MSR-VTT (50.9\%), LSMDC (25.8\%), and DiDeMo (47.2\%).
\end{abstract}

\section{Introduction}
\label{sec:intro}
\begin{figure}
    \begin{center}
        \includegraphics[width=1\linewidth]{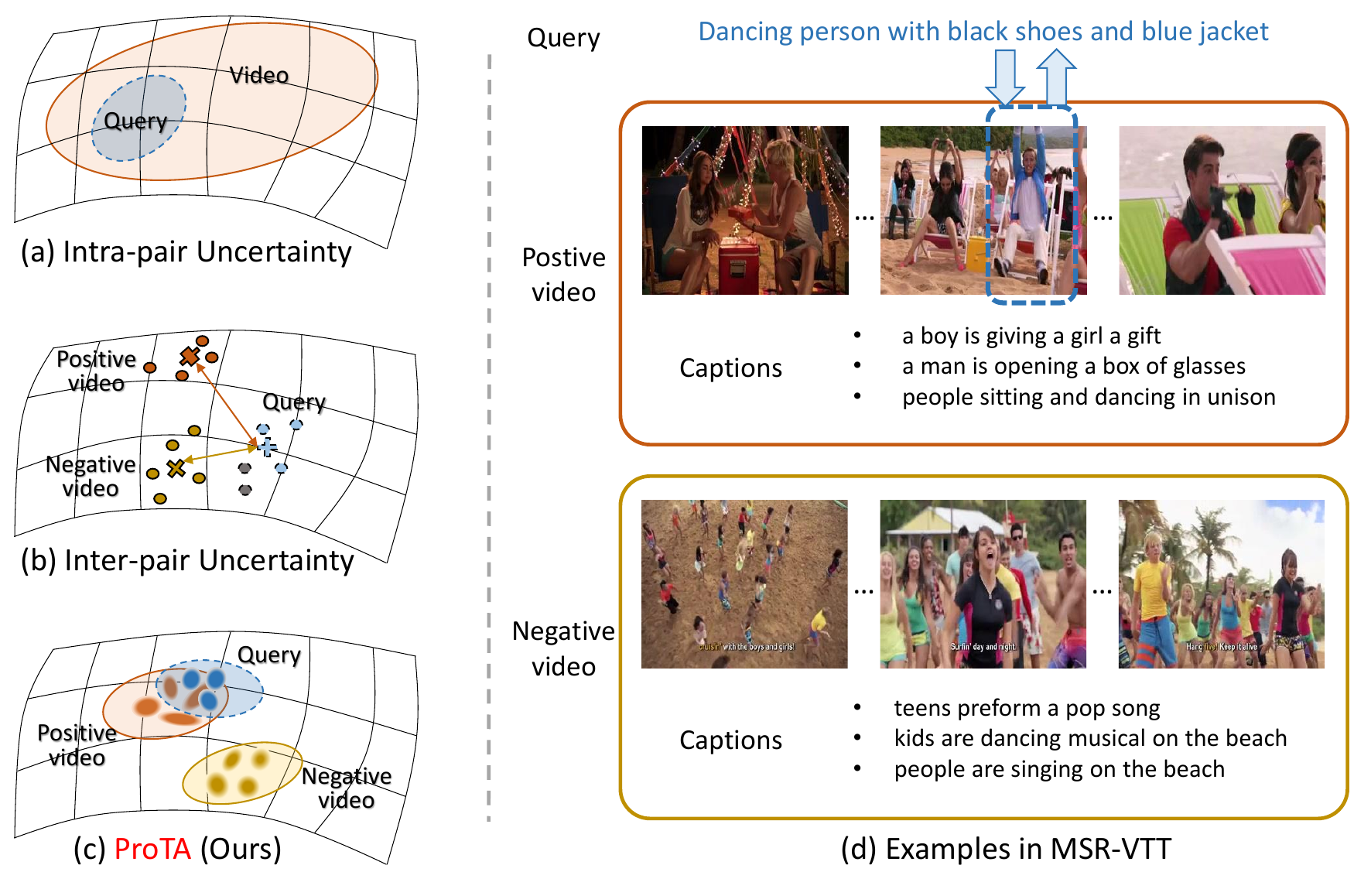}
    \end{center}
    \vspace{-0.5 cm}
    \caption{
(a) Intra-pair uncertainty. Videos often contain more information than their captions. (b) Inter-pair uncertainty. The query is easy to match negative videos with similar semantics. (c) ProTA can align cross-modality tokens at a fine-grained level and adaptively adjust the token-level distribution, which handles these two kinds of uncertainty in a unified manner. (d) Intra/inter-pair examples from MSR-VTT \cite{xu2016msr}.
    }
    \vspace{-0.32 cm}
    \label{introduction}  
\end{figure}

Text-video retrieval is an important cross-modal task for text-video semantics alignment, introducing more and more interest from the computer vision community \cite{cheng2021improving,luo2022clip4clip}. 
Most traditional methods fixed the pre-trained expert networks \cite{gabeur2020multi,liu2019use} to provide prior knowledge and focus on multi-modal interaction. However, the caption often describes the partial region of its corresponding video, and it is hard to match them due to content asymmetry.
Recently, transferring large-scale pre-training models to text-video retrieval has demonstrated its power. 
Several works aim to model the temporal relations \cite{fang2022transferring,gao2021clip2tv} and achieve cross-modality content alignment \cite{wang2021t2vlad}. These works employ point-based representations to describe the video or text features. However, there are some ambiguous words in captions, such as ``teens'' in Fig. \ref{introduction}(d). The word ``teens'' refers to ``boy'' or ``girl'' is unknown, and point-based representation lacks enough ability to describe them.

The above concerns are summarized into two kinds of data uncertainty:
1) \textbf{Intra-pair uncertainty.} The main content conveyed by each text-video pair is often partially related \cite{chun2021probabilistic}. As illustrated in Fig. \ref{introduction}(d), the caption only describes a short range of the original video. Selecting the most discriminative parts of different modalities and aligning them is not reasonable because they may be irrelevant. Besides, the redundant information in a video and the ambiguous words in a caption also hinder the content alignment. 
2) \textbf{Inter-pair uncertainty.} One video can be expressed through different texts, and one caption can describe some similar videos, which is easy to result in many false positive pairs. The point-based representation methods describe the feature as a fixed point in embedding space, which loses the representation diversity. 
Meanwhile, there are multiple understanding perspectives for the same video or text. Thus, the method reserving the representation diversity is more suited to text-video retrieval.

In this paper, we propose Probabilistic Token Aggregation (ProTA) to tackle intra-pair and inter-pair uncertainty simultaneously. Concretely, to solve the intra-pair uncertainty caused by content asymmetry, we propose Dual Partial-related Aggregation (DPA), including low-dimension and high-dimension aggregations. The low-dimension aggregation assigns different attention to each token. The high-dimension aggregation employs a Gram matrix with multiple radial basis functions to obtain intra-modality and inter-modality similarities. Then the video or text distributions are aggregated using a double-weighting strategy to capture partially related content.
Besides, we propose Token-based Probabilistic Alignment (TPA) to deal with inter-pair uncertainty. The token representation is formulated as a probabilistic distribution, and negative 2-Wasserstein distance is used to calculate the similarity between tokens.
Then dual partial-related aggregation and token-based probabilistic representation are gracefully combined. A KL regularization and an adaptive contrastive loss are introduced to keep feature representation diverse and pull the positive pair closer, which effectively handles the problem of inter-pair uncertainty.

Our main contributions are as follows: 
\textit{\textbf{(i)}} We propose a novel dual partial-related aggregation. The double-weighting strategy effectively handles the partially related cross-modality contents.
\textit{\textbf{(ii)}} We propose token-based probabilistic alignment to generate token-level probabilistic representation and adaptive margin with contrastive loss to enlarge the inter-pair distance.
\textit{\textbf{(iii)}}  The experimental results show our proposed model achieves significant improvements.

\section{Related Work}
\textbf{Video-text Retrieval}.
Early works often utilize offline feature extractors and focus on designing cross-modality fusion methods \cite{dzabraev2021mdmmt, gabeur2020multi}. Recent works have begun to employ an end-to-end manner to train their models. CLIP4Clip \cite{luo2022clip4clip} and CLIP2video \cite{fang2022transferring} transfer the prior knowledge from CLIP \cite{RN26} to video-text retrieval and propose a temporal modeling module. Following CLIP \cite{RN26}, these methods adopt infoNCE loss \cite{oord2018cpc} for optimization. Meanwhile, we follow the end-to-end manner and introduce a novel adaptive contrastive loss.

\textbf{Cross-modal Interaction}.
Early works mainly focus on improving the generalization ability of language and text encoders. They use a simple dot-product between the features.
CE \cite{liu2019use} employs the ASR and OCR information to extract more cues.  
To achieve fine-grained video-text retrieval,  CAMoE \cite{cheng2021improving} utilizes Mixture-of-Experts to obtain the action, scene, and object in each video and align them with the corresponding words. TACo \cite{yang2021taco} pre-defines the weight of tokens by computing inverse document frequency. 
However, these methods define hand-crafted concepts and attempt to align them. While we propose dual partial-related aggregation to achieve the data-driven and token-level interaction.

\textbf{Probabilistic Representation}.
Recent works have begun to employ the probabilistic feature to model the data/representation uncertainty. 
HIB \cite{RN22} models the feature embedding as a random variable and hedges the location of each input to a latent embedding space. PCME \cite{chun2021probabilistic} models image/text as a probability distribution and measure their distance by Monte-Carlo estimation. ProViCo \cite{RN29} supposes each clip has different distributions %
and can be embedded into the same probabilistic space. Instead, we model each token as a separate distribution, measuring the probabilistic similarity at the token level.

\begin{figure*}
    \begin{center}
        \includegraphics[width=1\linewidth]{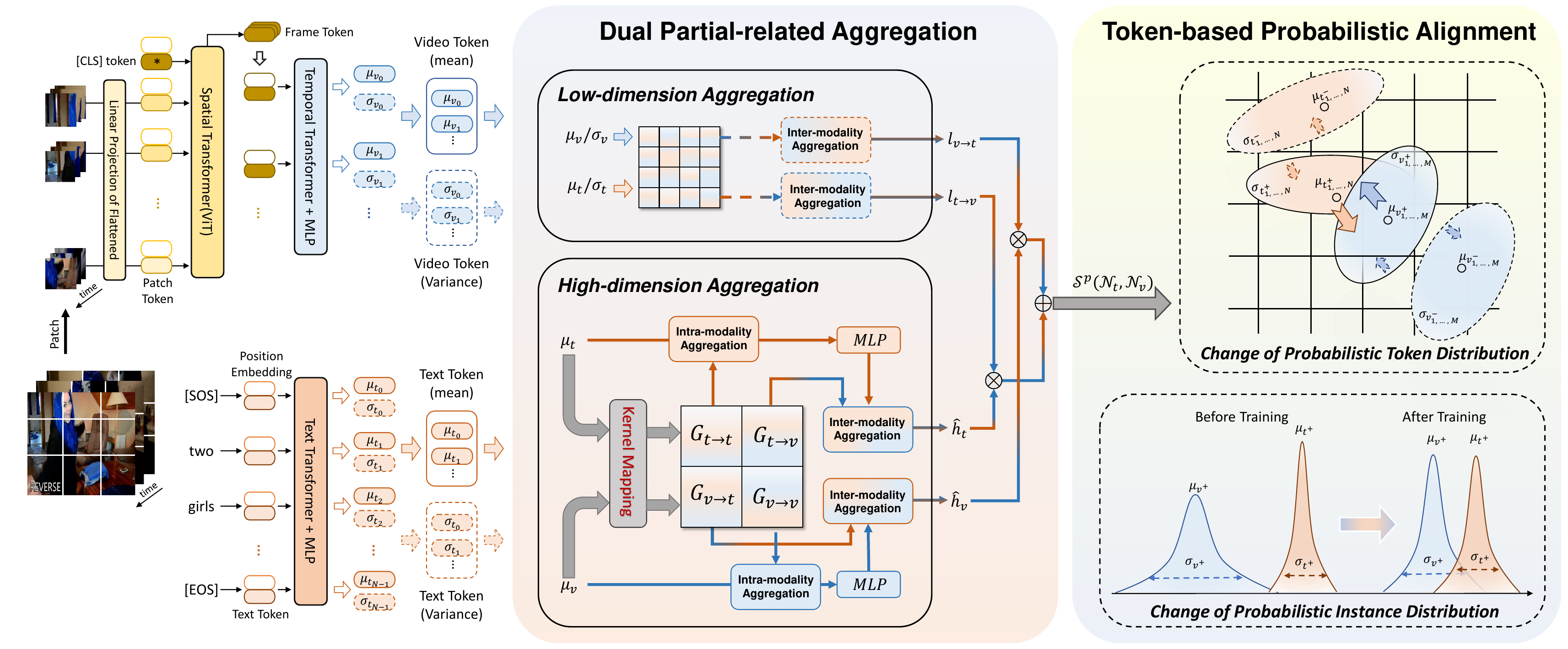}
    \end{center}
    \vspace{-4mm}
    \caption{
Overview of Probabilistic Token Aggregation. The two stream encoders are adopted to estimate the probabilistic distribution. We propose dual partial-related aggregation for token-level interaction, handling the problem of intra-pair uncertainty. The token-based probabilistic alignment is introduced to minimize the representation uncertainty, tackling inter-pair uncertainty.
    }
    \vspace{-0.22 cm}
    \label{framework}  
\end{figure*}

\section{Methodology}

The proposed Probabilistic Token Aggregation (ProTA) is illustrated in Fig. \ref{framework}. Each text is added with two special tokens \texttt{<SOS>} and \texttt{<EOS>}, which indicate the start and end of input text. Then a 12-layer Transformer from CLIP \cite{RN26} is employed to generate the text sequence tokens $f_t = \{t_{\texttt{SOS}}, t_1, \cdots, t_{N-2}, t_{\texttt{EOS}}\}$, where $N$ is the text token number. As for the video, we uniformly sample $M$ frames and adopt ViT to extract their spatial embeddings. Then these embeddings are processed by a temporal transformer \cite{luo2022clip4clip} to model the temporal correlation, outputting video token $f_v = \{v_0, v_1, \cdots, v_{M-1}\}$, where $M$ is the token number.

\subsection{Dual Partial-related Aggregation}
\label{section3.1}

\textbf{Low-dimension Aggregation.}
For the tokens of different modalities, token-level similarity matrix is formulated as: 
\begin{equation}
\begin{split}
\label{lowsim}
L_m =
\begin{bmatrix}
 s_{11} &  s_{12} &  \cdots \\
 s_{21} &  \ddots &  \cdots \\
 \vdots &  \vdots &  s_{t_iv_j} \\
 \vspace{-0.3 cm}
\end{bmatrix},
\end{split}
\end{equation}
where $s_{t_iv_j}$ is computed using dot product of $i^\text{th}$ text token and $j^\text{th}$ video token to represent their similarity.

To simplify the description, we only explain the text-to-video similarity computation process, and the video-to-text process has the same computation fashion. The cross-modality aggregation in low-dimension space is achieved by a weighted manner as follows:
\begin{equation}
\label{WS}
l_{t_i \rightarrow v} = \sum_{j=1}^{M} s_{t_i v_j} \frac{\exp{\lambda s_{t_i v_j}}}{\sum_{k=1}^{M}\exp{\lambda s_{t_i v_k }}}, \\
\end{equation}
where $i \in [1,\cdots, N]$, $\lambda$ is a temperature coefficient. The Eq. \ref{WS} is introduced to tackle the token-level content asymmetry, \eg a verb cannot be summarized by a frame, and a frame cannot be described by an adjective.
The vector $l_{t_i \rightarrow v}$ is the $i^\text{th}$ element of text-to-video similarity $l_{t\rightarrow v}$ in a low-dimension space.

\textbf{High-dimension Aggregation.}
Since video is a group of temporal scenarios, captions may merely describe the changes in a local region of videos. To further explore their partial-related content in multi-dimensional spaces, we propose to project the feature into multiple high-dimension spaces using radial basis functions (RBFs). In multiple high-dimension spaces, the intra-modality and inter-modality relationships are both explored and formulate a Gram matrix.

We explain the Gram matrix construction process as follows. The original text and video features, \textit{i.e.,} $f_t\in \mathbb{R}^{N\times d}$ and $f_{v}\in \mathbb{R}^{M \times d}$, are concatenated into a joint feature $f \in \mathbb{R}^{(N+M) \times d}$.
$R$ kernels are utilized for Gaussian mapping, which introduces multiple kinds of high-dimension embedding spaces. These generated high-dimension token-level similarities are averaged as follows: 
\begin{equation}
\begin{split}
\label{MSA}
g_{ij} = \frac{1}{R}\sum_{k=1}^{R} K_k(f_i, f_j), 
\vspace{-0.3 cm}
\end{split}
\end{equation}
where $f_i$ represents the $i^\text{th}$ vector in $f$ and $K_k$ indicates the $k^{th}$ kernel function.
All high-dimension token-level similarities $g_{ij}$ construct the Gram matrix $G\in \mathbb{R}^{(N+M)\times(N+M)}$, where $i$ and $j$ are in $[1,\cdots,N+M]$. 
With the help of multiple Gaussian kernel mapping, the Gram matrix disentangles the redundant similarity distribution in multiple high-dimension spaces and effectively captures the partially related content for alignment.
The text-to-text and video-to-video relations are dubbed $G_{t\rightarrow t}$ and $G_{v\rightarrow v}$ to represent the intra-modality similarities. $G_{t\rightarrow v}$ and $G_{v\rightarrow t}$ represent the inter-modality similarities. Therefore, Gram matrix $G$ can be expressed as:
\begin{equation}
\begin{split}
\label{MSA1}
G = 
\begin{bmatrix}
G_{t\rightarrow t}  & G_{t\rightarrow v} \\
G_{v\rightarrow t}  & G_{v\rightarrow v}\\
\end{bmatrix}. \\
\end{split}
\end{equation}

The intra-modality aggregation in high-dimension embedding space is processed as follows:
\begin{equation}
\begin{split}
\label{intraBlock}
h_{t\rightarrow t} = W_t(G_{t\rightarrow t}f_t),
\end{split}
\end{equation}
where $W_t$ is a two-layer MLP network. Intra-modality block matrix $G_{t\rightarrow t}$ is adopted to adjust the token-level weight, where redundant token representation is disentangled and re-aggregated to provide more attention to discriminative information.
Meanwhile, the inter-modality aggregation in high-dimension space is formulated as:
\begin{equation}
\begin{split}
\label{interBlock}
h_{t_i\rightarrow v} = \sum_{j=N+1}^{M+N} g_{t_i v_j} \frac{\exp{\xi g_{t_i v_j}}}{\sum_{k=1}^{M}\exp{\xi g_{t_i v_k}}}, \\
\end{split}
\end{equation}
where $i \in [1,\cdots, N]$, $\xi$ is a temperature coefficient, and $h_{t_i\rightarrow v}$ is the $i^\text{th}$ element in high-dimension text-to-video similarity $h_{t\rightarrow v}$. 
Rather than aligning discriminative textual representations, $h_{t\rightarrow v}$ is introduced to capture pair-relevant distribution at the cross-modal level.

By introducing Gram matrix with Gaussian mapping as an indicator, the re-weighted sequence tokens are formulated using intra-modality and inter-modality aggregation in high-dimension embedding space as follows:
\begin{equation}
\begin{split}
\label{}
\widehat{h_t}= h_{t \rightarrow v} \cdot h_{t\rightarrow t}.\\
\end{split}
\end{equation}

\textbf{Dual Partial-related Aggregation.}
The aggregations in low-dimension and high-dimension embedding spaces are combined, exploring more fine-grained relationships. Therefore, the similarity of each video-text pair is formulated as:
\begin{equation}
\begin{split}
\label{MSA2}
\mathcal{S}(f_t, f_v)=\frac{l_{t\rightarrow v} \cdot \widehat{h_t} + l_{v\rightarrow t} \cdot \widehat{h_v}}{2},
\end{split}
\end{equation}
where $l_{v\rightarrow t}$ and $\widehat{h_v}$ are obtained using the same manner as $l_{t\rightarrow v}$ and $\widehat{h_t}$.

\subsection{Token-based Probabilistic Alignment}
\label{section3.2}
\textbf{Token-based Probabilistic Representation.}
To deal with inter-pair uncertainty, we propose token-level probabilistic representation, while existing methods use the video/text-level representation.  
We adopt Gaussian distributions with mean $\mu$ and variance $\sigma$ to represent each token. For text sequence tokens, we parameterize distribution as:
\begin{equation}
\begin{split}
\label{mean}
\mu_t=W^\text{mean}(f_t), \quad & \sigma^2_t=\exp W^\text{var}(f_t), 
\end{split}
\end{equation}
where $W^\text{mean}$ and $W^\text{var}$ represent the MLP networks. The mean $\mu_t$ and variance $\sigma_t^2$ have the same dimension as $f_{t}$, \textit{i.e.,} $\mathbb{R}^{N\times d}$, which represent mean feature and representation uncertainty of token sequences. The mean $\mu_{v}$ and variance $\sigma_v$ of video are obtained using the same equation as Eq. \ref{mean}.
In this way, the token-level feature is represented by a probabilistic distribution $\mathcal{N}(\mu, \sigma)$, which introduces the representation diversity. For example, `person' in a caption can represent `man', `woman', or some other kinds, while the original feature only can represent a specific/certain one.

Moreover, we propose using the 2-Wasserstein distance to measure the token-level distance first and utilize dual partial-related aggregation to obtain the probabilistic similarity of the video-text pair. The probabilistic token similarity is calculated using negative 2-Wasserstein distance as follows:
\begin{equation}
\label{MSA3}
s_{t_i v_j}^{p} =-\sqrt{ \Vert \mu_{t}^{i} -\mu_{v}^j \Vert^2 + \Vert \sigma_t^{i^2} - \sigma_v^{j^2} \Vert^2}.
\end{equation} 
Then probabilistic similarity $s_{t_i v_j}^{p}$ replaces the corresponding $s_{t_i v_j}$ in Eq. \ref{lowsim} and generates the probabilistic low-dimension similarity matrix $L_m^{p}$. Then, the low-dimension text-to-video similarity $l_{t \rightarrow v}^p$ can be obtained using Eq. \ref{WS} in the same manner. 
Since mean $\mu_t$ and $\mu_v$ contain most of the semantics, we utilize them to construct high-dimension Gram matrix $G^p$ and obtain $\widehat{h_t^p}$ and $\widehat{h_v^p}$ with the same operation as high-dimension aggregation. Therefore, the probabilistic similarity of each video-text pair is aggregated as:
\begin{equation}
\label{MSA4}
\mathcal{S}^p(\mathcal{N}_t, \mathcal{N}_v)=\frac{l_{t\rightarrow v}^p \cdot \widehat{h_t^p} + l_{v\rightarrow t}^p \cdot \widehat{h_v^p}}{2}.
\end{equation}



To 
prevent $\delta^2$ 
degrading into point representations, a Kullback-Liebler (KL) regularization is adopted to keep the feature distribution in accordance with the normal distribution $\mathcal{N}(0, I)$ as:
\begin{equation}
\begin{split}
\label{MSA5}
L^\text{kl}=\frac{1}{2}[&\text{KL}(\mathcal{N}(\mu_t, \sigma_t) || \mathcal{N}(0, I)) + \\ &\text{KL}(\mathcal{N}(\mu_v, \sigma_v) || \mathcal{N}(0, I))].
\end{split}
\end{equation}

\textbf{Adaptive Contrastive Loss.}
For model training, we propose an adaptive margin with infoNCE \cite{oord2018representation} to supervise the proposed framework. We only present the text-to-video loss function to simplify the description as follows:
\begin{equation}
\label{MSA6}
L_{t2v} = - \mathop{\mathbb{E}}\limits_{t\in b_i} \log \frac{\exp[{(\mathcal{S}^p(\mathcal{N}_t, \mathcal{N}_v^+)-\eta \cdot m_{t2v})/\tau]}}{\sum \exp[{\mathcal{S}^d(\mathcal{N}_t, \mathcal{N}_v)/\tau]}},\\
\end{equation}
where $b_i$ represents the $i^\text{th}$ mini-batch, $\mathcal{N}_v^+$ is the positive distribution in each mini-batch, $\tau$ is the temperature coefficient, $\eta$ is a hyper-parameter, $m_{t2v}$ is the adaptive margin and generated using the following equation:
\begin{equation}
\begin{split}
\label{KL}
m_{t2v} = \exp [-\text{KL}(\mathcal{N}(\mu_t, \sigma_t) || \mathcal{N}(\mu_v, \sigma_v))],\\
\end{split}
\end{equation}
where the KL divergence plays the role of indicator to adjust penalty dynamically. 
For example, when a text-video pair has a close KL divergence distance, the margin penalty $m_{t2v}$ will increase, which gives the model a harder learning target and results in a discriminative model. 
The whole contrastive loss is illustrated as:
\begin{equation}
\begin{split}
\label{MSA7}
L^\text{contra} = \frac{1}{2}(L_{t2v} + L_{v2t}),
\end{split}
\end{equation}
where $L_{v2t}$ is obtained in the same manner as $L_{t2v}$. 
For model training, the total objective of the proposed Probabilistic Token Aggregation (ProTA) is formulated as:
\begin{equation}
\begin{split}
\label{MSA8}
L^{all}= L^\text{contra} + \beta L^\text{kl}, 
\end{split}
\end{equation}
where $\beta$ is the weight to control trade-off.

\section{Experiments}
\subsection{Experimental Setting}
\textbf{Datasets.}
\emph{MSR-VTT-9k} \cite{xu2016msr} includes 10,000 videos, where each video contains 20 captions. we report results on the 1k-A split (9000 for training, 1000 for testing).
\emph{LSMDC} \cite{rohrbach2017movie} is composed of 118,081 videos, where 1,000 videos are selected for evaluation. \emph{DiDeMo} \cite{anne2017localizing} includes over 10k videos. Following the same setting as \cite{lei2021less,liu2021hit,luo2022clip4clip}, we  concatenate all descriptions of a video as a caption and treat this dataset as a video-paragraph retrieval task.

\textbf{Evaluation Metric.} Following \cite{miech2018learning}, we  report Recall at rank K (R@K), median rank (MdR), and mean rank (MnR) as metrics, where the higher R@K, and lower MdR and MnR represent the better performance.
 
\textbf{Implementation Details.} 
We follow the same experimental setting of prior work \cite{fang2022transferring}
to initialize parameters for both datasets. Besides, $R$, $\lambda$ and  $\xi$ for relation-aware aggregation are set to 5, 100, and 100. We choose $\eta$ and $\beta$ as 5e-4 in uncertainty-based alignment. We use the AdamW optimizer and set the batch size to 128. The learning rate is 1e-7 for frame and text encoder, and 1e-3 for the proposed modules.  More details are listed in the supplemental material.

\begin{table}
		\caption{Performance on MSR-VTT-9k \cite{xu2016msr}. `LdA': Low-dimension aggregation. `intra': intra-modality aggregation. `inter': inter-modality aggregation. `$G$': Gram matrix with Gaussian kernel.  `AC': adaptive contrastive loss. `$G^\dagger$': Gram matrix using dot product. `$\S$': probabilistic similarity. }
	\centering
	\resizebox{0.48\textwidth}{!}{
		\begin{tabular}{c|ccc|ccc} 
	\hline
    \multicolumn{1}{c}{}&\multicolumn{3}{c}{Text $\Longrightarrow$  Video} &\multicolumn{3}{c}{Video $\Longrightarrow$ Text} \\
    \hline
	Method & R@1 & R@5 & R@10 & R@1 & R@5 & R@10 \\
    \hline
Global Pooling \cite{luo2022clip4clip}   & 44.5 & 71.4 & 81.6 & 42.7 & 70.9 & 80.6 \\
Mean-Max \cite{wang2022disentangled} & 46.3 & 73.7 & 83.2 & 44.0 & 73.5 & 82.1 \\ \hline
LdA & 46.5 & 74.0 & 83.4 & 44.2 & 74.1 & 82.6 \\
LdA+$G^\dagger_\text{intra}$ & 46.2 & 74.2 & 83.3 & 44.0 & 73.9 & 83.0 \\
LdA+$G^\dagger_\text{intra+inter}$ & 46.7 & 74.3 & 83.7 & 44.3 & 74.1 & 83.3 \\
LdA+$G_\text{intra}$ & 46.8 & 74.5 & 83.9  & 44.2 & 74.4 & 83.6 \\
LdA+$G_\text{intra+inter}$ (DPA) & 47.0 & 75.0 & 84.1 & 44.6 & 74.6 & 84.0\\
\hline
DPA$^\S$  & 46.7&	74.4&	83.2&	43.9&	73.4&	83.8 \\
DPA$^\S$+$L^\text{kl}$  & 47.7&	75.1&	83.8&	45.4&	74.8&	84.0 \\
DPA$^\S$+$L^\text{kl}$+AC (ProTA) &\textbf{48.1}	&\textbf{75.4} &\textbf{84.3}& \textbf{45.9} &\textbf{75.5}	&\textbf{84.6} \\
	\hline
	\end{tabular}}
	\label{PAI_total}
\end{table}

\subsection{Ablation Studies}
\label{Ablation Experiments}
\textbf{Effectiveness of DPA.} 
In Tab. \ref{PAI_total}, compared with the global based method \cite{luo2022clip4clip}, Mean-Max \cite{wang2022disentangled} brings improvements by providing fine-grained tokens. However, these methods adjusted weights within one modality, which did not consider the alignment of partially related content. Instead, we propose Low-dimension Aggregation (LdA) that assigns different weights based on token distribution.
Then to disentangle the redundant information and re-aggregate partially related content in multiple high-dimension spaces, we adopt the Gram matrix with multiple radial basis functions as the indicator `$G$'. Based on `LdA', we utilize the intra-modality blocks `$G_\text{intra}$' to enhance the weight of the discriminative token. We add inter-blocks `$G_\text{inter}$' to generate `LdA+$G_\text{intra+inter}$' and introduce aligned relations, which achieves better performance.  Besides, we show the results of Gram matrix ($G^\dagger$) using dot product. Without kernel mapping for high-dimension aggregation, $G^\dagger$ brings limited improvement.

\begin{table}[t]
\caption{Text-to-video results on MSR-VTT-9k \cite{xu2016msr}.}
	\label{MSRVTTTable}
	\begin{center}
		\resizebox{0.45\textwidth}{!}{
			\begin{tabular}{c|ccccc} 
    \hline
	Method & R@1 & R@5 & R@10 & MdR & MnR\\
    \hline
	FROZEN \cite{bain2021frozen}            & 31.0 & 59.5 & 70.5	& 3.0  & - \\
	HIT \cite{liu2021hit}         & 30.7 & 60.9 & 73.2 & 2.6  & -     \\
	MDMMT \cite{dzabraev2021mdmmt}           & 38.9 & 69.0 & 79.7	& 2.0  & 16.5 \\
    CLIP4Clip \cite{luo2022clip4clip} & 44.5 & 71.4 & 81.6	& 2.0  & 15.3 \\
    CLIP2Video \cite{fang2022transferring} & 45.6 & 72.6 & 81.7	& 2.0  & 14.6 \\
    X-pool \cite{gorti2022x}  & 46.9 & 72.8 & 82.2	& 2.0  & 14.3 \\
    TS2Net \cite{liu2022ts2} & 47.0 & 74.5 & 83.8	& 2.0  & 13.0 \\
    \rowcolor{gray!20} \textbf{ProTA (Ours)} & \textbf{48.1} & \textbf{75.4} & \textbf{84.3}	& \textbf{2.0}  & \textbf{12.5}  \\
	\hline
    CLIP2TV + ViT-B/16 \cite{gao2021clip2tv}& 48.3 & 74.6 & 82.8 & 2.0 & 14.9 \\
    CenterCLIP + ViT-B/16 \cite{zhao2022centerclip} & 48.4 & 73.8 & 82.0 & 2.0 & 13.8 \\
	TS2Net+ ViT-B/16  \cite{liu2022ts2} & 49.4 & 75.6 & 85.3 & 2.0  & 13.5 \\
	\rowcolor{gray!20} \textbf{ProTA (Ours)} + ViT-B/16 & \textbf{50.9} & \textbf{77.0} & \textbf{85.4} & \textbf{1.0}  & \textbf{11.1}  \\
	\hline
		\end{tabular}}
	\end{center}
	\vspace{-0.1 cm}
	
\end{table}

\begin{figure}[t]
    \begin{center}
        \includegraphics[width=1\linewidth]{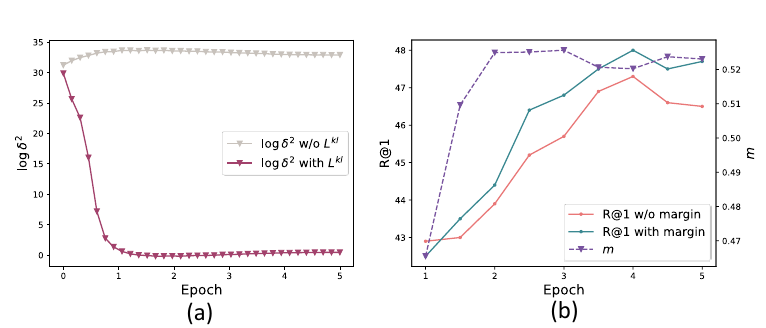}
    \end{center}
\vspace{-0.6 cm}
    \caption{
  \textbf{(a)} Analysis of $L_{kl}$ by training with MSR-VTT-9k \cite{xu2016msr}.  \textbf{(b)} Adaptive margin versus performance in MSR-VTT-9k 1kA test, where $m$ represents average of $m_{t2v} + m_{v2t}$.}
    \label{loss}  
\end{figure}

 \textbf{Effectiveness of  TPA.}
The results are shown in the last four lines of Tab. \ref{PAI_total}. Dual partial-related aggregation is extended to measure the distribution distance (`DPA$^\S$'). The KL regularization is added to keep the token-level distribution diversity, which generates `DPA$^\S$+$L^\text{KL}$'. In Fig. \ref{loss}(a), we also depict the uncertainty change represented by $\log\sigma_t^2$. As depicted, the KL regularization keeps the distribution to a proper range, where the representation uncertainty is decreased. In addition, applying the adaptive margin in the loss function introduces further improvement (`DPA$^\S$+$L^\text{KL}$+AC'). In Figure  \ref{loss}(b), we illustrate the change of model performance. As the training progresses, the model ends its warm-up, and the effect of the adaptive margin appears gradually. The model with adaptive margin performs better than the one without margin.

\begin{table}
		\caption{Text-to-video results on LSMDC \cite{rohrbach2017movie} testing set. }
	\begin{center}
		\resizebox{0.45\textwidth}{!}{
			\begin{tabular}{c|ccccc} 
    \hline
	Method & R@1 & R@5 & R@10 & MdR & MnR \\
    \hline
MMT \cite{gabeur2020multi} & 12.9 & 29.9 & 40.1 & 19.3 &65.0  \\
FROZEN \cite{bain2021frozen}                  & 15.0 & 30.8 & 39.8	  & 20.0  & -    \\
TeachText \cite{croitoru2021teachtext} & 17.2 & 36.5 & 46.3 & 13.7 & -  \\
MDMMT \cite{dzabraev2021mdmmt}  & 18.8 & 38.5 &47.9 &12.3 & 58.0    \\
CLIP4Clip \cite{luo2022clip4clip}   & 22.6 & 41.0 & 49.1 & 11.0 & 61.0 \\
TS2-Net \cite{liu2022ts2} & 23.4 & 42.3 & 50.9 & 9.0 & 56.9   \\
X-Pool \cite{gorti2022x} & 25.2 & 43.7 & 53.5 & 8.0 & 53.2   \\
\hline
\rowcolor{gray!20} \textbf{ProTA (Ours)} & \textbf{25.8} & \textbf{47.3} & \textbf{57.7} & \textbf{7.0} & \textbf{45.3} \\
	\hline
		\end{tabular}}
	\end{center}
		\vspace{-0.1 cm}

	\label{LSMDC}
\end{table}

\begin{table}
		\caption{Text-to-video results on DiDeMo \cite{anne2017localizing} testing set.}
	\begin{center}
		\resizebox{0.45\textwidth}{!}{
			\begin{tabular}{c|ccccc} 
    \hline
	Method & R@1 & R@5 & R@10 & MdR & MnR \\
    \hline
ClipBERT \cite{lei2021less} & 20.4 & 48.0& 60.8 & 6.0 & - \\
TeachText \cite{croitoru2021teachtext} & 21.1 & 47.3 & 61.1 & 6.3 & -  \\
FROZEN \cite{bain2021frozen}           & 31.0 & 59.8 & 72.4	  & 3.0  & -    \\
CLIP4Clip \cite{luo2022clip4clip}      & 43.4 & 70.2 & 80.6 & 2.0 & 17.5 \\
TS2-Net \cite{liu2022ts2}              & 41.8 & 71.6 & 82.0 & 2.0 & 14.8   \\
\hline
\rowcolor{gray!20} \textbf{ProTA (Ours)} & \textbf{47.2} & \textbf{74.6} & \textbf{83.0} & \textbf{2.0} & \textbf{11.6} \\
\hline
		\end{tabular}}
	\end{center}
		\vspace{-0.1 cm}

	\label{DiDemo}
\end{table}

\subsection{Comparisons with State-of-the-art Methods}
We compare with other state-of-the-art methods on three datasets. The text-to-video results are reported in Tab. \ref{MSRVTTTable}, \ref{LSMDC}, and \ref{DiDemo}. As observed, without any augmentation \cite{cheng2021improving, wu2022cap4video}, our method still achieves significant progress, which improves R@1 by at least \textbf{2.0\%} across different evaluation metrics. 
Moreover, we observe that the improvement is more remarkable in Tab. \ref{DiDemo}.
We adopt the same token length as other state-of-the-art methods, which use a longer token length. Prominent results represent that ProTA works well on cross-modal pairs with diverse content and prove its generalization. 
Besides, we also conduct the experiments adopting ViT-B/16 backbone and report results in the bottom of Tab. \ref{MSRVTTTable}, which outperforms the previous best method by a large margin of +1.5$\%$ in terms of text-to-video R@1 metric. \textbf{The video-to-text results can be found in the supplementary material.}

\subsection{Qualitative Comparisons.}
 \begin{figure}
    \begin{center}
        \includegraphics[width=1\linewidth]{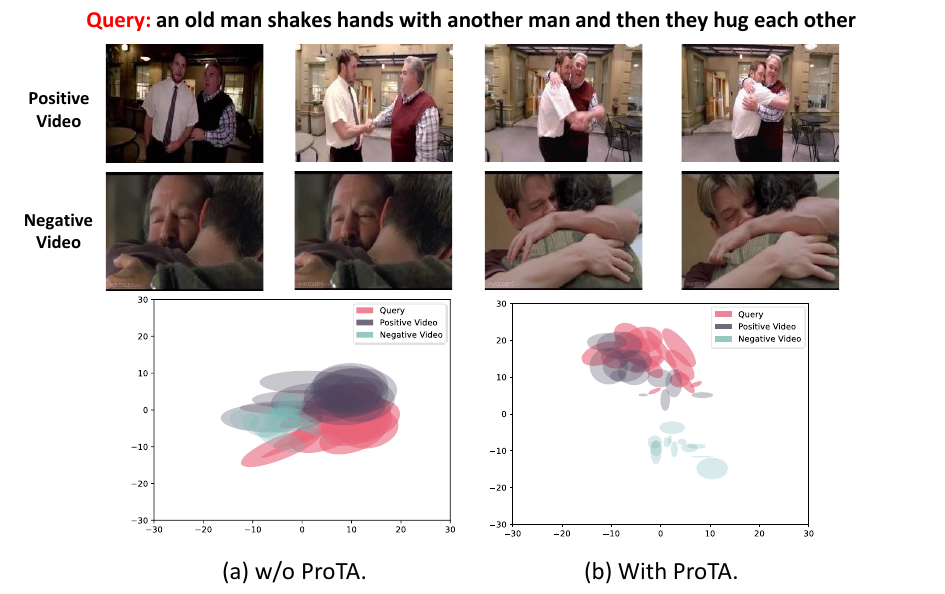}
    \end{center}
    \vspace{-0.4 cm}
    \caption{
    Visualization of probabilistic distribution.}
    \vspace{-0.42 cm}
    \label{novel2}  
\end{figure}

In Fig. \ref{novel2}, we visualize the features to verify the effectiveness of proposed ProTA. Concretely, we select text-to-video results from MSR-VTT, where the negative video is wrongly retrieved without DPA and TPA. We adopt t-SNE \cite{donahue2014decaf} to depict their distributions, where each ellipse represents one token distribution. With limited textual descriptions and a lot of scenario changes in the video, these pairs contain much uncertainty. Without ProTA, the negative pairs easily overlap each other, causing the wrong matching. After introducing ProTA, the distance between the positive pair decreases, and the distance between the negative pair increases, while the token uncertainty is significantly decreased.

\section{Conclusion}
This paper  proposes a novel method called Probabilistic Token Aggregation (ProTA). 
In this work, ProTA introduces dual partial-related aggregation to distinguish cross-modal pairs in multi-dimension space, alleviating the intra-pair uncertainty.
Furthermore, token-based probabilistic alignment is employed to represent token as probabilistic distribution, reducing the inter-pair uncertainty in the token level.

\footnotesize
\bibliographystyle{IEEEbib}
\bibliography{icme2024}

\clearpage
\appendix

In this supplementary material, we present detailed information including: 1. Introduction of used datasets and evaluation protocols; 2. Implementation Details. 3. More video-to-text comparisons with state-of-the-art methods;  4. Ablation studies;  5. More qualitative results. 
\begin{table*}
	\caption{Retrieval performance comparison trained on MSR-VTT-9k \cite{xu2016msr} and evaluated on 1k-A test.}
	\begin{center}
		\resizebox{0.8\textwidth}{!}{
			\begin{tabular}{c|ccccc|ccccc} 
				\hline
				\multicolumn{1}{c}{}& \multicolumn{5}{c}{Text $\Longrightarrow$ Video} &\multicolumn{5}{c}{Video $\Longrightarrow$ Text} \\
				\hline
				Method & R@1 & R@5 & R@10 & MdR & MnR & R@1 & R@5 & R@10 & MdR & MnR\\
				\hline
				CE \cite{liu2019use} & 20.9 & 48.8 & 62.4 & 6.0 & 28.2 & 20.6 & 50.3 & 64.0 & 5.3 & 25.1 \\
				MMT \cite{gabeur2020multi}    & 26.6 & 57.1 & 69.6	& 4.0  & 24.0 & 27.0 & 57.5 & 69.7 & 3.7 & 21.3 \\
				SUPPORT-SET \cite{patrick2020support}    & 27.4 & 56.3 & 67.7 & 3.0  & -    & 26.6 & 55.1 & 67.5 & 3.0 & - \\
				T2VLAD \cite{wang2021t2vlad}             & 29.5 & 59.0 & 70.1 & 4.0  & -    & 31.8 & 60.0 & 71.1 & 3.0 & -  \\
				FROZEN \cite{bain2021frozen}            & 31.0 & 59.5 & 70.5	& 3.0  & -    & -    &    - & -    &   - & - \\
				CLIPforward \cite{portillo2021straightforward} & 31.2 & 53.7 & 64.2 & 4.0  & -    & 27.2 & 51.7 & 62.6 & 5.0 & - \\
				HIT \cite{liu2021hit}         & 30.7 & 60.9 & 73.2 & 2.6  & -    & 32.1 & 62.7 & 74.1 & 3.0 & - \\
				MDMMT \cite{dzabraev2021mdmmt}           & 38.9 & 69.0 & 79.7	& 2.0  & 16.5 & -    &    - & -    &   - & - \\
				CLIP4Clip \cite{luo2022clip4clip} & 44.5 & 71.4 & 81.6	& 2.0  & 15.3 & 42.7 & 70.9 & 80.6 & 2.0 & 11.6 \\
				CLIP2Video \cite{fang2022transferring} & 45.6 & 72.6 & 81.7	& 2.0  & 14.6 & 43.5 & 72.3 & 82.1 & 2.0 & 10.2 \\
				X-pool \cite{gorti2022x}  & 46.9 & 72.8 & 82.2	& 2.0  & 14.3 & - & - & - & - & - \\
				TS2Net \cite{liu2022ts2} & 47.0 & 74.5 & 83.8	& 2.0  & 13.0 & 45.3 & 74.1 & 83.7 & 2.0 & 9.2 \\
				\rowcolor{gray!20}ProTA (Ours) & \textbf{48.1} & \textbf{75.4} & \textbf{84.3}	& \textbf{2.0}  & \textbf{12.5} & \textbf{45.9} & \textbf{75.5} & \textbf{84.6} & \textbf{2.0} & \textbf{9.0} \\
				\hline
				CLIP2TV + ViT-B/16 \cite{gao2021clip2tv}& 48.3 & 74.6 & 82.8 & 2.0 & 14.9 & 46.5 & 75.4 & 84.9 & 2.0 & 10.2 \\
				CenterCLIP + ViT-B/16 \cite{zhao2022centerclip} & 48.4 & 73.8 & 82.0 & 2.0 & 13.8 & 47.7 & 75.0 & 83.3 & 2.0 & 10.2 \\
				TS2Net+ ViT-B/16  \cite{liu2022ts2} & 49.4 & 75.6 & 85.3	& 2.0  & 13.5 & 46.6 & 75.9 & 84.9 & 2.0 & 8.9 \\
				\rowcolor{gray!20}ProTA (Ours) + ViT-B/16 & \textbf{50.9} & \textbf{77.0} & \textbf{85.4} & \textbf{1.0}  & \textbf{11.1}  & \textbf{48.5} & \textbf{77.0} & \textbf{87.0} & \textbf{2.0} & \textbf{7.9} \\
				\hline
		\end{tabular}}
	\end{center}
	\label{MSRVTTTable}
\end{table*}

\section{Datasets}
\begin{itemize}
	\item  \textbf{MSR-VTT} \cite{xu2016msr} contains 10,000 videos, where each video contains 20 captions. We adopt two protocols to evaluate our proposed method. In the main paper, we report results trained on MSR-VTT-9k and test in 1k-A protocol. In supplementary material, we demonstrate the results of the full protocol. Specifically,  full protocol \cite{dzabraev2021mdmmt,luo2020univilm} is the standard split which includes 6,513 videos for train, 497 videos for validation, and 2,990 videos for testing. In this protocol, each video contains multiple independent captions, which are all used in text-to-video retrieval. When reporting the results of video-to-text retrieval, we adopt the maximum similarity among all corresponding captions for a given video query. 
	\item \textbf{LSMDC} \cite{rohrbach2017movie} is composed of 118,081 videos collected from 202 movies. Each video has only one caption. There are about 100k videos in the training set and 1k videos are utilized as the test set.
	\item \textbf{DiDeMo} \cite{anne2017localizing} is the dataset with dense localization by natural language queries, which is proposed for temporal video grounding. Following \cite{lei2021less,liu2021hit,luo2022clip4clip}, we utilize this dataset for text-video retrieval, concatenate all descriptions of a video as a caption and treat this dataset as a video-paragraph retrieval task. The dataset contains over 10k videos, which adopts 8,395 videos for training, 1,065 videos for validation, and 1,004 for testing. \textbf{Note that the methods we compare in the paper and supplementary material follow the same setting, including the same length of video and text tokens and video-paragraph retrieval task.}
\end{itemize}

\section{Implementation Details.} 
The frame encoder and text encoder are initialized by CLIP (ViT-B/32). We reuse parameters of CLIP to initialize temporal transformer. The layer, head, and width of spatial and temporal transformer in video encoder is 12, 8, 768, and 4, 8, 512, respectively. The text encoder is a 12-layer transformer with 8 attention heads and 768 widths. To match the representation of different modalities, two linear projections are utilized to encode them into dimension of 512. The length of video and caption is 12 and 32 for MSR-VTT \cite{xu2016msr} and LSMDC \cite{rohrbach2017movie}. As for DiDeMo \cite{anne2017localizing}, we set the length of video and caption to 64, which is the same setting as existing methods \cite{liu2019use, zhang2018cross}.
$R$, $\lambda$ and  $\xi$ for relation-aware aggregation are set to 5, 100, and 100. We choose $\eta$ and $\beta$ as 5e-4 in uncertainty-based alignment.
We fine-tune our model with Adam optimizer with cosine schedule decay. The batch size is 128.  The training epoch is 5 for MSR-VTT \cite{xu2016msr} and LSMDC \cite{rohrbach2017movie}, and  10 for DiDeMo \cite{anne2017localizing}. The learning rate is 1e-7 for frame encoder and text encoder, and 1e-3 for the proposed modules.

\begin{table}
	\caption{Video-to-text retrieval results on the LSMDC \cite{rohrbach2017movie} testing set. }
	\begin{center}
		\resizebox{0.45\textwidth}{!}{
			\begin{tabular}{c|ccccc} 
				
				\hline
				Method & R@1 & R@5 & R@10 & MdR & MnR \\
				\hline
				JSFusion \cite{yu2018joint} & 12.3 & 28.6 & 38.9 & 20.0 & - \\
				CLIPforward \cite{portillo2021straightforward}        & 6.8 & 16.4 & 22.1 & 73.0   & - \\
				CLIP4Clip \cite{luo2022clip4clip}   & 20.8 & 39.0 & 48.6 & 12.0 & 54.2 \\
				\hline
				\rowcolor{gray!20} \textbf{ProTA (Ours)} & \textbf{24.5} & \textbf{46.8} & \textbf{56.6} & \textbf{7.0} & \textbf{41.7} \\
				\hline
		\end{tabular}}
	\end{center}
	\label{LSMDC1}
\end{table}

\begin{table}
	\caption{Video-to-text results on the DiDeMo \cite{anne2017localizing} testing set. }
	\begin{center}
		\resizebox{0.45\textwidth}{!}{
			\begin{tabular}{c|ccccc} 
				
				\hline
				Method & R@1 & R@5 & R@10 & MdR & MnR \\
				\hline
				CE \cite{liu2019use}                    & 15.6 & 40.9 & - & 8.2 & 42.4   \\
				TeachText \cite{croitoru2021teachtext}  & 21.1 & 47.3 & 61.1 & 6.3 & - \\
				CLIP4Clip \cite{luo2022clip4clip}      & 42.5 & 70.6 & 80.2 & 2.0 & 11.6 \\
				\hline
				\rowcolor{gray!20} \textbf{ProTA (Ours)} & \textbf{46.2} & \textbf{73.2} & \textbf{83.4} & \textbf{2.0} & \textbf{9.6} \\
				\hline
		\end{tabular}}
	\end{center}
	\label{didemo}
\end{table}

\begin{table}
	\caption{Video-to-text retrieval results on the MSR-VTT-full \cite{xu2016msr}.}
	\begin{center}
		\resizebox{0.45\textwidth}{!}{
			\begin{tabular}{c|ccccc} 
				\hline
				Method & R@1 & R@5 & R@10 & MdR & MnR \\
				\hline
				CE \cite{liu2019use} & 15.6 & 40.9 & 55.2 & 8.2 & 38.1 \\
				T2VLAD \cite{wang2021t2vlad} & 20.7 & 48.9 & 62.1 & 6.0 \\
				CLIPforward \cite{portillo2021straightforward} & 40.3 & 69.7 & 79.2 & 2.0 & - \\
				CLIP4Clip \cite{luo2022clip4clip}  & 53.8 & 80.9 & 88.6 & 1.0 & 7.4 \\
				CLIP2Video \cite{fang2022transferring}& 54.6 & 82.1 & 90.8 & 1.0 & 5.3 \\
				\hline
				\rowcolor{gray!20} \textbf{ProTA (Ours)} & \textbf{55.3} & \textbf{83.2} & \textbf{91.3} & \textbf{1.0} & \textbf{4.8} \\
				\hline
		\end{tabular}}
	\end{center}
	\label{msrvttfull1}
\end{table}

\begin{table}
	\caption{Text-to-video retrieval results on the MSR-VTT-full \cite{xu2016msr}.}
	\begin{center}
		\resizebox{0.45\textwidth}{!}{
			\begin{tabular}{c|ccccc} 
				\hline
				Method & R@1 & R@5 & R@10 & MdR & MnR \\
				\hline
				CE \cite{liu2019use} & 10.0 & 29.0 &41.2 & 16.0 & 86.8 \\
				T2VLAD \cite{wang2021t2vlad} & 12.7 & 34.8 & 47.1 & 12.0 & - \\
				CLIPfowrard \cite{portillo2021straightforward} & 21.4 & 41.1 & 50.4 & 10.0& - \\
				CLIP4Clip \cite{luo2022clip4clip}  & 28.9 & 54.1 & 64.8 & 4.0 & 48.4 \\
				CLIP2Video \cite{fang2022transferring}& 29.8 & 55.5 & 66.2 & 4.0 & 45.5 \\
				
				\hline
				\rowcolor{gray!20} \textbf{ProTA (Ours)} & \textbf{31.1} & \textbf{56.4} & \textbf{67.4} & \textbf{4.0} & \textbf{44.7} \\
				\hline
		\end{tabular}}
	\end{center}
	\label{msrvttfull2}
\end{table}

\section{Comparisons with State-of-the-art Methods} 
We compare our proposed ProTA with state-of-the-art methods on video-to-text benchmarks including MSR-VTT-9k \cite{xu2016msr},  LSMDC \cite{rohrbach2017movie} and DiDemo \cite{anne2017localizing} The detailed results are demonstrated in Tab. \ref{MSRVTTTable},  \ref{LSMDC1} and \ref{didemo}. Note that since MSR-VTT-9k is the most commonly used data set, we demonstrate all the text-to-video and video-to-text results in Tab. \ref{MSRVTTTable}, where some text-to-video results are also presented in the main paper.  To further demonstrate the effectiveness and generalization ability of our method, we also show the results of the full protocol in MSR-VTT. The full protocols require the model to find videos based on multiple corresponding captions, while the maximum similarity among all corresponding captions is needed for a given video query, increasing the difficulty of matching. In Tab. \ref{msrvttfull1} and \ref{msrvttfull2}, our proposed method still shows priority compared with others.

\begin{table}
	\caption{Comparison results with different settings of $\lambda$ and $\xi$ on MSR-VTT-9k \cite{xu2016msr}. Note that we report the results based on dual partial-related aggregation without probabilistic alignment.
	}
	\begin{center}
		\resizebox{0.48\textwidth}{!}{
			\begin{tabular}{cc|ccc|ccc} 
				\hline
				\multicolumn{2}{c}{}&\multicolumn{3}{c}{Text $\Longrightarrow$  Video} &\multicolumn{3}{c}{Video $\Longrightarrow$ Text} \\
				\hline
				$\lambda $&$\xi$ & R@1 & R@5 & R@10 & R@1 & R@5 & R@10 \\
				\hline
				\textit{max}  &-                   & 46.3 & 74.0 & 83.4 & 44.0 & 74.0 & 82.8 \\
				200  &-                   & 46.5 & 74.2 & 83.5 & 44.0 & 74.1 & 83.2 \\
				\rowcolor{gray!20}100 &- & \textbf{46.8} & \textbf{74.5} & \textbf{83.9} & \textbf{44.2} &	\textbf{74.4} & \textbf{83.6} \\
				50 &-                     &46.7  & 74.2 & 82.7 & 43.8 &	73.7 & 82.9 \\
				\hline
				100   &\textit{max}                &46.8	&74.8 &83.8& 44.3 &74.5 &83.7 \\
				100   &200                          &\textbf{47.1}	&\textbf{75.4} &83.7& 44.3 &74.3 &83.8 \\
				\rowcolor{gray!20}100      &100    &47.0	&75.0 &\textbf{84.1}& \textbf{44.6} &\textbf{74.6} &\textbf{84.0} \\
				100   &50                           &46.5	&74.8 &83.7& 44.0 &74.0 &83.5 \\
				\hline
		\end{tabular}}
	\end{center}
	\label{prob_matching}
\end{table}

\begin{table}
	\caption{Performance comparison with different numbers of Gaussian kernels on MSR-VTT-9k \cite{xu2016msr}.}
	\begin{center}
		\resizebox{0.48\textwidth}{!}{
			\begin{tabular}{c|ccc|ccc} 
				\hline
				\multicolumn{1}{c}{}&\multicolumn{3}{c}{Text $\Longrightarrow$  Video} &\multicolumn{3}{c}{Video $\Longrightarrow$ Text} \\
				\hline
				Method & R@1 & R@5 & R@10 & R@1 & R@5 & R@10 \\
				\hline
				R=1 & 46.7 & 74.6 & 83.9 & 44.2 & 74.2 & 83.5 \\
				R=3 & 46.8 & 74.9 & 84.0 & 44.4 & 74.4 & 83.8 \\
				\rowcolor{gray!20}R=5 & 47.0 & 75.0 & 84.1 & 44.6 & 74.6 & 84.0 \\
				R=7 & 47.0 & 75.1 & 84.3 & 43.6 & 74.7 & 83.6 \\
				R=9 & 46.9 & 74.7 & 83.7 & 43.4 & 74.2 & 83.2 \\
				\hline
		\end{tabular}}
	\end{center}
	
	\label{PAI_kernel}
\end{table}

\begin{figure*}
	\begin{center}
		\includegraphics[width=1\linewidth]{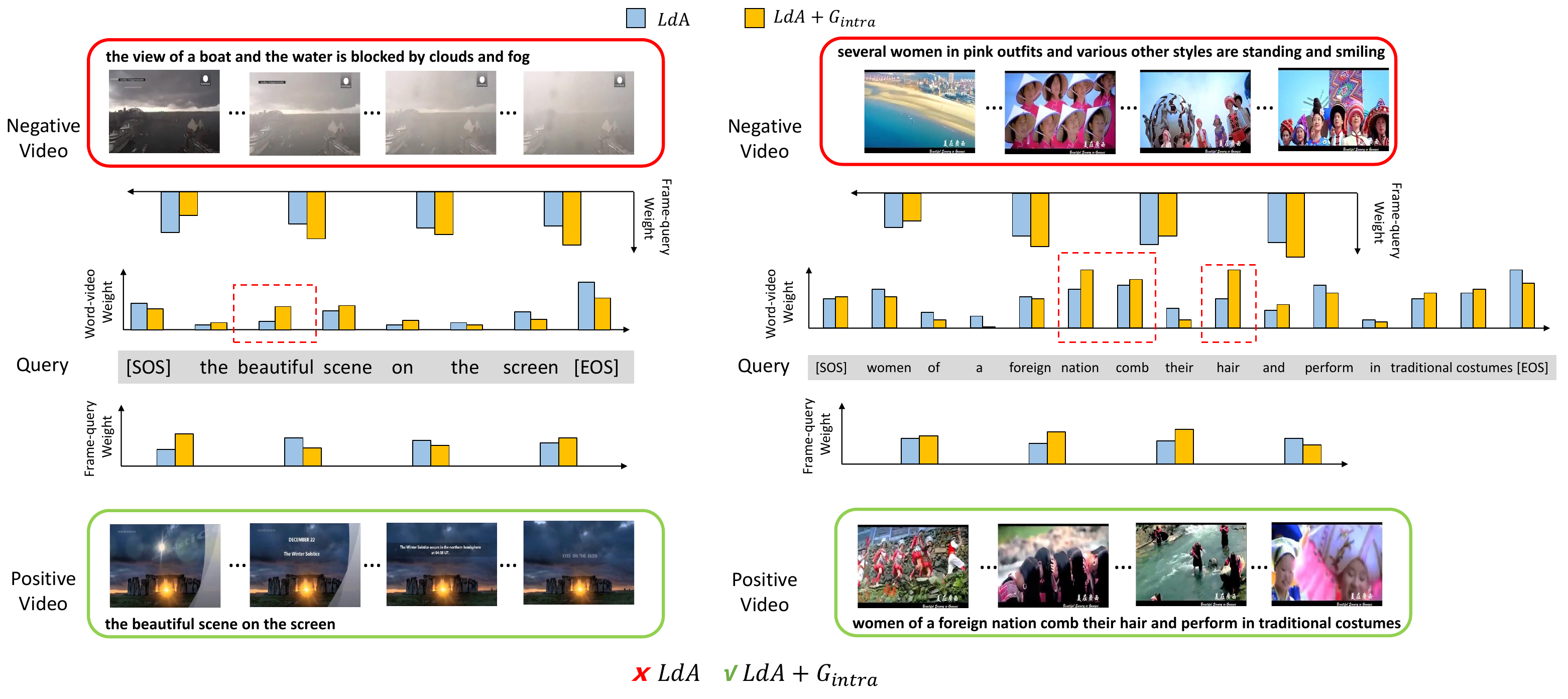}
	\end{center}
	\vspace{-0.4cm}
	\caption{
		Ablation of intra-modality aggregation. We adopt the same sampling strategy as paper and select 4 frames uniformly from frame sequences.}
	\label{novle1_intra}  
\end{figure*}

\begin{figure*}
	\begin{center}
		\includegraphics[width=1\linewidth]{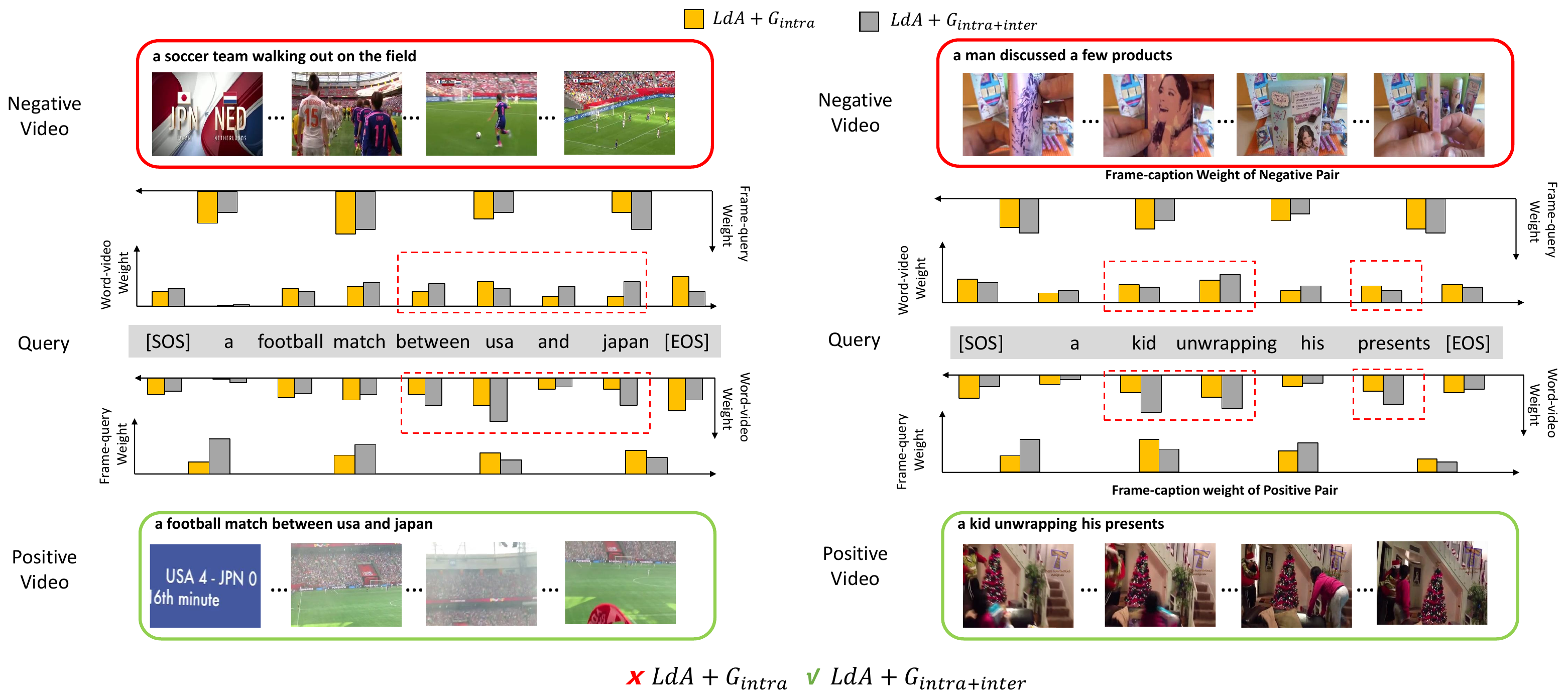}
	\end{center}
	\vspace{-0.4cm}
	\caption{
		Ablation of inter-modality aggregation. We adopt the same sampling strategy as paper and select 4 frames uniformly from frame sequences. }
	\label{novle1_inter}  
\end{figure*}

\section{Ablation studies}

\subsection{Temperature in DPA}
We propose different temperature coefficients of dual partial-related aggregation (DPA) in Eq. \textcolor{blue}{2} and Eq. \textcolor{blue}{6} in the main paper to balance the weight of every token towards the corresponding modality. Only considering the similarity of the most important token, the model is easily disturbed by a few tokens, ignoring other discriminative frames and words. Meanwhile, it is not suitable to utilize a single frame to represent the word with temporal descriptions like the verb, and vice versa. Therefore, we adopt a weighted similarity with temperature coefficient $\lambda$ and $\xi$ for low-dimension and high-dimension aggregation respectively. The results of different settings are reported in Tab. \ref{prob_matching}. We choose 100 for $\lambda$ and 100 for $\xi$, which empirically yields the best overall performance.

\textbf{Number of Kernels.}
In Tab. \ref{PAI_kernel}, we conduct comparisons to explore the effects of kernel numbers. The performance of dual partial-related aggregation is improved as the kernel number increases. Employing several kernels for mappings, the redundant semantics is well disentangled and re-aggregated in multiple high-dimensional spaces to capture partially related content. Besides, when the number is too large, the mapping generates more noisy information due to the limited representation in low-dimension space.

\begin{figure}
	\begin{center}
		\includegraphics[width=1\linewidth]{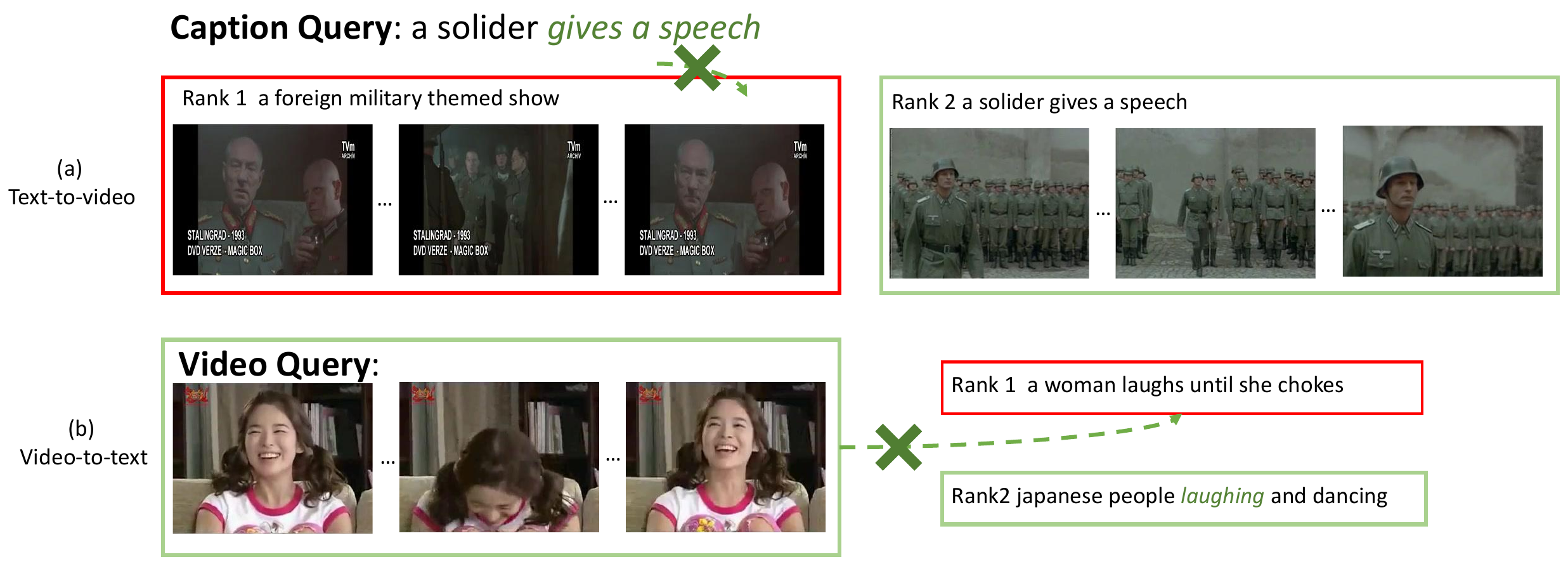}
	\end{center}
	\caption{
		Failure cases of text-to-video retrieval (a) and video-to-text retrieval (b)  on MSR-VTT-9k.  }
	\label{failure_case}  
\end{figure}

\section{More Qualitative Results}
We visualize more samples to prove the effectiveness of adjusting token weight, where Fig. \ref{novle1_intra} depicts the ablation of intra-modality aggregation and Fig. \ref{novle1_inter} represents the ablation of inter-modality aggregation. In Fig. \ref{novle1_intra}, the token weight is re-weighted within the modality. The model learns to focus more on distinguishing pairs according to detailed descriptions like "beautiful", "nation" and "hair". In Fig. \ref{novle1_inter}, we demonstrate the cases that adding inter-modal aggregation, where the text-to-video weight is determined by both relations of word tokens and aligned frame tokens. As observed, the weight of word token is adjusted according to different scenarios, \eg, the importance of "usa" is larger when matching with the positive pairs, and the first frame in the positive video is highlighted as the most discriminated scenario. By introducing the inter-modality relations, the weight of tokens can be dynamically adjusted to capture the most relevant representations in pairs.

In Fig. \ref{failure_case}, we demonstrate the failure cases of text-to-video retrieval and video-to-text retrieval, where our model fails to retrieve the corresponding cross-modal pair at the top. However, we find that the most relevant videos and text are actually retrieved. The failed reason may come from ambiguous annotations such as ‘gives a speech’ and ‘laughing’. For example, the video query in Fig. \ref{failure_case}(b) can be well described by the two captions, due to the polysemous description without specific details.

\end{document}